\documentclass[12pt]{article}
\usepackage{amsfonts}
\usepackage{a4wide}
\usepackage{amsmath,amscd,amsthm,a4,amssymb}
\usepackage{graphicx}

\begin{document}

\begin{center}
{\Large \textbf{On the Kolmogorov neural networks}}

\bigskip

\bigskip

\large{Aysu Ismayilova} \large{and Vugar E. Ismailov}\footnote{
Corresponding author at: The Institute of Mathematics and Mechanics,
9 B. Vahabzadeh str., AZ1141, Baku, Azerbaijan; E-mail:
vugaris@mail.ru}

\end{center}

\bigskip

\bigskip

\textbf{Abstract.} In this paper, we show that the Kolmogorov two hidden
layer neural network model with a continuous, discontinuous bounded or
unbounded activation function in the second hidden layer can precisely
represent continuous, discontinuous bounded and all unbounded multivariate
functions, respectively.

\bigskip



\bigskip

\bigskip

\begin{center}
{\large \textbf{1. Introduction}}
\end{center}

There are a lot of papers in neural network literature on the capability of
special neural networks, called the Kolmogorov neural networks or
Kolmogorov's mapping neural networks, to precisely represent each continuous
multivariate function. But precise representation in other function classes
has not been considered there. In this paper, we show that the Kolmogorov
networks have an extreme power of representing not only continuous, but also
discontinuous bounded and all unbounded multivariate functions.

The idea of constructing of the above mentioned neural networks stems from
the famous Kolmogorov superposition theorem. This theorem positively solves
Hilbert's 13th problem. Hilbert in his address to the International Congress
of Mathematicians held in Paris in 1900, outlined 23 outstanding
mathematical problems, the 13th of which asked: Is the root of the equation
\begin{equation*}
x^{7}+ax^{3}+bx^{2}+cx+1=0
\end{equation*}%
a superposition of continuous functions of two variables? Hilbert thought
the answer should be negative -- surely functions of three variables are
more complex than those of two. It should be remarked that this problem
resisted all efforts to prove it for more than 50 years. Almost all the
mathematicians, interested in this problem, were attempting to prove the
validity of Hilbert's conjecture. But in 1957, Kolmogorov \cite{Kol} refuted
all expectations by proving that each continuous function of three and more
variables can be represented by superpositions of continuous functions of
one variable and the single function of two variables, namely the addition
function $x+y$.

In literature, there are many versions of Kolmogorov's superposition
theorem. From the perspective of applications in neural networks, we cite
the following version, which is due to Sprecher \cite{Spr2,Spr3}:

\bigskip

\textbf{Theorem 1.1.} \textit{Let$~d\geq 2$ and $\gamma \geq 2d+2$ be given
integers and $\mathbb{I}=[0,1]$. There exists a universal monotonic
increasing function $\varphi $ of the class $Lip[\ln 2/\ln \gamma ]$ such
that every continuous $d$-variable function $f:\mathbb{I}^{d}\rightarrow
\mathbb{R}$ has the representation}

\begin{equation*}
f(x_{1},...x_{d})=\sum_{q=0}^{2d}g_{q}\left( \sum_{p=1}^{d}\lambda
_{p}\varphi (x_{p}+aq)\right), \eqno(1.1)
\end{equation*}%
\textit{where $g_{q}$ is some continuous one-variable function depending on $%
f.$ Here $a=\left[ \gamma (\gamma -1)\right] ^{-1},$ $\lambda _{1}=1,\lambda
_{p}=\sum_{r=1}^{\infty }\gamma ^{-(p-1)(d^{r}-1)/(d-1)}$ for $p=2,3,...,d$.}

\bigskip

Eq. (1.1) has the following interpretation as a feedforward neural network
consisting of an input layer, two hidden layers and an output layer. The
input layer having $d$ neurons sends signals $x_{1},...x_{d}$ to the first
hidden layer with $d(2d+1)$ neurons and the activation function $\varphi $.
The $(q,p)$-th neuron $y_{q,p}$ ($0\leq q\leq 2d,$ $1\leq p\leq d$) produces
the signal $\varphi (x_{p}+aq)$. These signals are sent to the second hidden
layer consisting of $2d+1$ neurons and the activation functions $g_{q}$. The
$q$-th neuron $z_{q}$ ($0\leq q\leq 2d$) produces the signal $g_{q}\left(
\sum_{p=1}^{d}\lambda _{p}y_{q,p}\right) $. Finally, the output layer with
the unique output neuron just sums up these last signals to produce the
number $f(x_{1},...x_{d})$. Feedforward neural networks with this structure
are usually called the Kolmogorov neural networks. Figure 1 displays
Kolmogorov's neural network in case of $d=2$.

\begin{figure}[ht]
\hfill
\par
\begin{center}
\includegraphics[width=1.00\textwidth]{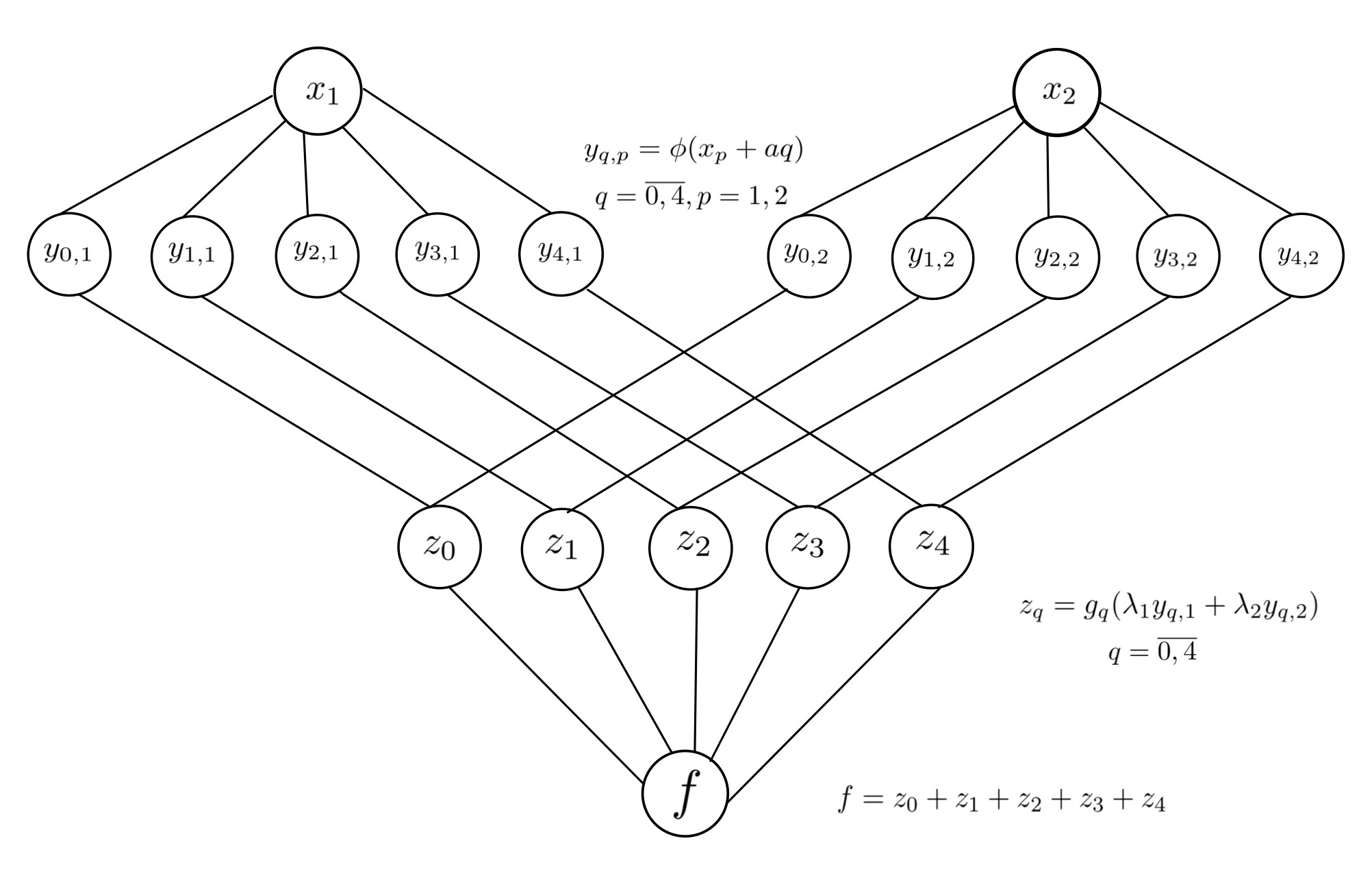}
\end{center}
\caption{The neural network interpretation of Kolmogorov's superposition
theorem.}
\end{figure}

The relevance of Kolmogorov's superposition theorem to neural networks was
first observed by Hecht-Nielsen \cite{Hec}. In Hecht-Nielsen's
interpretation the Kolmogorov neural network (called in \cite{Hec}
Kolmogorov's mapping neural network) had three layers, two hidden layers in
Fig. 1 were identified as a single layer. Although Kolmogorov's theorem
reveals the profound character of feedforward neural networks to precisely
represent each continuous function, it was considered by some authors as
non-constructive. For example, Girosi and Poggio pointed out that
Kolmogorov's network is not useful. In \cite{Gir} they argued that for an
implementation of a network that has good properties, the functions
corresponding to the layers in the network have to be smooth, which is not
the case for the functions $\varphi $ and $g_{q}$ in Kolmogorov's network.
This criticism was addressed by K\r{u}rkova \cite{Kur,Kur1} pointing out
that the relevance of Kolmogorov's superposition theorem to approximation by
neural networks is different. K\r{u}rkova substituted the precise
representation with an approximation of the target function $f$. For this
purpose she used sigmoidal functions $\sigma $ (which is defined as a
function with the property $\lim_{t\rightarrow -\infty }\sigma (t)=0$ and $%
\lim_{t\rightarrow +\infty }\sigma (t)=1$) and finite linear combinations $%
\sum_{k}c_{k}\sigma (w_{k}x-\theta _{k})$ to approximate functions of one
variable, in particular Kolmogorov's inner universal and outer functions. As
a result, the number of terms in the outer sum was increased, but her method
enabled an estimation of the number of hidden neurons depending on the
approximation accuracy $\varepsilon $. Note that in \cite{Kur1} the number
of hidden neurons increases when the approximation accuracy $\varepsilon
\rightarrow \infty $.

The above dependency of $\varepsilon $ on the number of neurons was
eliminated in Nakamura, Mines, and Kreinovich \cite{Nak}. They developed
algorithms that generate the activation functions with guaranteed accuracy
and keep number of hidden neurons independent of $\varepsilon $. All
operations and functions in \cite{Nak} were defined constructively, which
means that they are implementable in a computer program. However, as noted
by the authors of \cite{Nak}, the algorithms that they constructed in the
proofs are very complicated and not suitable for practical usage. For other
approximative, but constructive approaches to function approximation by
using Kolmogorov's superposition theorem, see \cite{Fig,Ige,Nee}.

Using K\r{u}rkova's ideas, Sprecher and Katsura \cite{Kat} constructed a
sequence of functions $\left\{ \varphi _{n}\right\} _{n=1}^{\infty }$ that
converges to the inner function $\varphi $ and constructed $2d+1$ series of
functions that converge to the outer functions $g_{q}$ in (1.1).

Later Sprecher \cite{Spr2,Spr3} developed a numerical algorithm for the
computation of the inner and outer functions in (1.1). In these papers, the
inner function $\varphi $ is defined as an extension of a function which is
explicitely defined on the set of so-called terminating rational numbers.
Note that these numbers are dense in $\mathbb{R}$. He proved continuity and
monotonicity of the resulting function $\varphi $.

The papers \cite{Spr2,Spr3} were discussed in K\"{o}ppen \cite{Kop}, where
it was pointed out that Sprecher's function $\varphi $ does not possess the
continuity and monotonicity properties. To fill this gap, K\"{o}ppen
suggested a modified inner function and stated its continuity. He defined $%
\varphi $ recursively on the same set of terminating rational numbers $%
\mathcal{D}$ and claimed that this recursion terminates. K\"{o}ppen assumed
that there exists an extension from $\mathcal{D}$ to $\mathbb{R}$ as in
Sprecher's construction and that this extended $\varphi $ is monotone
increasing and continuous, but he did not give a proof for it. Such a proof
was given in Braun and Griebel \cite{Bra} and Braun \cite{Bra1}. That is, it
was shown that K\"{o}ppen's $\varphi $ indeed exists, i.e., it is well
defined and has the necessary continuity and monotonicity properties.

\bigskip

\bigskip

\begin{center}
{\large \textbf{2. Main result}}
\end{center}

In this section we show that the above Theorem 1.1 can be generalized to
discontinuous bounded and also to all unbounded functions. In addition, we
prove that in all cases the outer functions $g_{q}$ can be replaced by a
single outer function $g$.

Obviously, all functions on a given compact set $X$\ can be divided into the
following three nonintersecting classes. These are the classes of
continuous, discontinuous bounded and unbounded functions. Note that if a
function is continuous on $X$, then it is automatically bounded. The
following theorem gives a precise representation formula for each of these
classes.

\bigskip

\textbf{Theorem 2.1.} \textit{Assume$~d\geq 2$ and $\gamma \geq 2d+2$ are
given integers and $\mathbb{I}=[0,1]$. Set $a=\left[ \gamma (\gamma -1)%
\right] ^{-1},$ $\lambda _{1}=1,\lambda _{p}=\sum_{r=1}^{\infty }\gamma
^{-(p-1)(d^{r}-1)/(d-1)}$ for $p=2,3,...,d$ and $b_{q}=(2d+1)q$ for $%
q=0,1,...,2d$. Then there exists a universal monotonic increasing function $%
\varphi $ of the class $Lip[\ln 2/\ln \gamma ]$ with the property:}

\textit{Each $d$-variable function $f:\mathbb{I}^{d}\rightarrow \mathbb{R}$
can be precisely represented in the form}

\begin{equation*}
f(x_{1},...x_{d})=\sum_{q=0}^{2d}g\left( \sum_{p=1}^{d}\lambda _{p}\varphi
(x_{p}+aq)+b_{q}\right),\eqno(2.1)
\end{equation*}%
\textit{where $g$ is a one-variable function depending on $f$. If $f$ is
continuous, then $g$ can be chosen continuous as well. If $f$ is
discontinuous bounded, then $g$ is discontinuous bounded; and if $f$ is
unbounded, then $g$ is unbounded.}

\bigskip

Theorem 2.1 gives rise to the following feedforward neural network model.
This model contains four layers: the input layer with $d$ neurons $%
x_{1},...x_{d}$, the first hidden layer with $d(2d+1)$ neurons $y_{q,p}$, $%
0\leq q\leq 2d,$ $1\leq p\leq d$, the second hidden layer with $2d+1$
neurons $z_{q}$, $0\leq q\leq 2d$, and the output layer with a single neuron
$w$. Activation functions of the first and second hidden layers are $\varphi
$ and $g$, respectively. The connecting rules between the layers are as
follows:

\begin{eqnarray*}
\text{The input layer} &\text{:}&\text{ }x_{1},...x_{d}; \\
\text{The first hidden layer} &\text{:}&\text{ }y_{q,p}=\varphi (x_{p}+aq)%
\text{ for }q=0,1,...,2d;~p=1,...,d; \\
\text{The second hidden layer} &\text{:}&\text{ }z_{q}=g\left(
\sum_{p=1}^{d}\lambda _{p}y_{q,p}+b_{q}\right) \text{ for }q=0,1,...,2d; \\
\text{The output layer} &\text{:}&\text{ }w=\sum_{q=0}^{2d}z_{q}.
\end{eqnarray*}

Theorem 2.1 means that any multivariate function $f(x_{1},...x_{d})$ can be
implemented by such a network. That is, $w=f(x_{1},...x_{d})$. The only
parameter depending on $f$ is the activation function $g$ of the second
hidden layer. It carries continuity and boundedness properties of the given $%
f$. More precisely, $g$ is continuous if $f$ is continuous, $g$ is
discontinuous bounded if $f$ is discontinuous bounded, and $g$ is unbounded
if $f$ is unbounded.

\bigskip

\textbf{Proof.} Using Sprecher's result, it is easy to prove (2.1) for
continuous $f:\mathbb{I}^{d}\rightarrow \mathbb{R}$. Note that for such a
function representation (1.1) is valid. In the proof of (1.1) $\varphi $ was
constructed in such a way that $\varphi (\mathbb{I})=\mathbb{I}$ and $%
\varphi (x+1)=\varphi (x)+1$ for $x\in \mathbb{I}$ (see \cite{Spr2}). This
means that $\varphi ([0,2])=[0,2].$ Since $0\leq x_{p}+aq<2$ for all indices
$p$ and $q$, we have $0\leq \varphi (x_{p}+aq)<2.$ On the other hand it is
not difficult to check that $\max \left\{ \lambda _{1},...,\lambda
_{p}\right\} =1.$ Hence the ranges of all the functions $\sum_{p=1}^{d}%
\lambda _{p}\varphi (x_{p}+aq) $ in (2.1) fall into $[0,2d]$. It follows
that the ranges of the functions

\begin{equation*}
\Psi _{q}(x_{1},...x_{d}):=\sum_{p=1}^{d}\lambda _{p}\varphi
(x_{p}+aq)+b_{q},~~q=0,1,...,2d,
\end{equation*}%
are pairwise disjoint.

Let $Y_{q}$ denote the range of $\Psi _{q}(x_{1},...x_{d})$. That is,

\begin{equation*}
Y_{q}:=\Psi _{q}(\mathbb{I}^{d})\text{ for all }q=0,1,...,2d.
\end{equation*}%
Set
\begin{equation*}
Y:=\cup _{q=0}^{2d}Y_{q}.
\end{equation*}%
Note that all $Y_{q}$ and $Y$ are compact sets. Construct the function $g$
on $Y$ by the following way:
\begin{equation*}
g(y):=g_{q}(y-b_{q})\text{ if }y\in Y_{q},\text{ }q=0,1,...,2d.\eqno(2.2)
\end{equation*}%
Since $Y_{i}\cap Y_{j}\neq \emptyset $, for all $0\leq i,j\leq 2d$, $i\neq j$%
, this formula makes $g$ a well-defined function on $Y.$ Clearly, $g$ is
continuous on $Y$ and we can extend $g$ by continuity to the whole $\mathbb{R%
}$. In fact, there are many ways of doing this. Now taking (2.2) into
account in (1.1), we obtain (2.1).

\bigskip

Now let us prove (2.1) for bounded $f:\mathbb{I}^{d}\rightarrow \mathbb{R}$.
We use the above proved fact that representation (2.1) holds for continuous
multivariate functions defined on $\mathbb{I}^{d}$. Let for any compact set $%
X $, $C(X)$ and $B(X)$ stand for the spaces of continuous functions on $X$
and bounded functions on $X$, respectively.

Consider the operator

\begin{equation*}
T:C(Y)\rightarrow C(\mathbb{I}^{d}),\text{ }Tg=\sum_{q=0}^{2d}g\left(
\sum_{p=1}^{d}\lambda _{p}\varphi (x_{p}+aq)+b_{q}\right) .\eqno(2.3)
\end{equation*}

Since (2.1) is valid for all $f\in C(\mathbb{I}^{d})$, the operator $T$ in
(2.3) is a surjection. Consider also the dual spaces $C(Y)^{\ast }$ and $C(%
\mathbb{I}^{d})^{\ast }.$ These are the spaces of regular real-valued
measures of finite total variation defined on Borel subsets of $Y$ and $%
\mathbb{I}^{d}$, respectively.

The conjugate operator $T^{\ast }:C(\mathbb{I}^{d})^{\ast }\rightarrow
C(Y)^{\ast }$ has the form

\begin{equation*}
T^{\ast }\mu =\sum_{q=0}^{2d}\Psi _{q}\circ \mu .
\end{equation*}%
Here $\Psi _{q}\circ \mu $ is a measure in $C(Y_{q})^{\ast }$, which is
defined as follows

\begin{equation*}
\left[ \Psi _{q}\circ \mu \right] (Q)=\mu (\Psi _{q}^{-1}(Q)),\text{ for all
}Q\in \mathcal{B}_{q}\text{,}
\end{equation*}%
where $\mathcal{B}_{q}$ is the set of Borel subsets of $Y_{q}.$

It is a well known fact in Functional Analysis that an operator $%
F:U\rightarrow V$ between Banach spaces $U$ and $V$ is a surjection if and
only if the conjugate operator $F^{\ast }:V^{\ast }\rightarrow U^{\ast }$ is
one-to-one and closed (and vice versa). The last is equivalent to the
inequality
\begin{equation*}
\left\Vert x\right\Vert \leq \epsilon \left\Vert F^{\ast }x\right\Vert \text{%
,}
\end{equation*}%
for all $x\in V^{\ast }$ and some $\epsilon >0$ (see, e.g., \cite[Ch. 4]{Rud}%
). Applying this fact to our problem, we obtain that there exists a positive
number $\epsilon $ such that the inequality

\begin{equation*}
\left\Vert \mu \right\Vert \leq \epsilon \left\Vert T^{\ast }\mu \right\Vert
=\epsilon \left\Vert \sum_{q=0}^{2d}\Psi _{q}\circ \mu \right\Vert \leq
\epsilon \sum_{q=0}^{2d}\left\Vert \Psi _{q}\circ \mu \right\Vert \eqno(2.4)
\end{equation*}%
holds for all $\mu \in C(\mathbb{I}^{d})^{\ast }$.

For any compact set $X$, consider a linear space $l_{1}(X)$ consisting of
discrete measures in $X$. That is, $\nu \in l_{1}(X)$ means that

\begin{equation*}
\nu =\sum_{i=1}^{\infty }a_{i}\delta _{t_{i}},\text{ }\left\Vert \nu
\right\Vert =\sum_{i=1}^{\infty }\left\vert a_{i}\right\vert <\infty
\end{equation*}%
where $\{a_{i}\}_{i=1}^{\infty }$ is a sequence of real numbers such that $%
\sum_{i=1}^{\infty }\left\vert a_{i}\right\vert <\infty ,$ $%
\{t_{i}\}_{i=1}^{\infty }$ is a sequence in $\mathbb{I}^{d}$ and $\delta
_{t_{i}}$ are points masses at $t_{i}.$

Note that $l_{1}(X)\subset C(X)^{\ast }$. Thus (2.4) holds also for all $\mu
\in l_{1}(\mathbb{I}^{d}).$

Now construct the following Banach space

\begin{equation*}
S=\left\{ \nu =(\nu _{0},...,\nu _{2d}):\nu _{q}\in l_{1}(Y_{q}),~0\leq
q\leq 2d\right\} ,\text{ }\left\Vert \nu \right\Vert
=\sum_{q=0}^{2d}\left\Vert \nu _{q}\right\Vert .
\end{equation*}

Consider its dual $S^{\ast }$. Since for any $X$, the dual of $l_{1}(X)$ is
the space of bounded functions on $X$ (see, e.g., \cite[Ch. 4]{Dun}), the
space $S^{\ast }$ has the following structure:

\begin{equation*}
S^{\ast }=\left\{ g=(g_{0},...,g_{2d}):g_{q}\in B(Y_{q}),~0\leq q\leq
2d\right\} ,\text{ }\left\Vert g\right\Vert =\max_{0\leq q\leq 2d}\left\Vert
g_{q}\right\Vert .
\end{equation*}%
Note that $g\in S^{\ast }$ acts on $\nu \in S$ as follows

\begin{equation*}
g[\nu ]=\sum_{q=0}^{2d}\int_{Y_{q}}g_{q}d\nu _{q}.
\end{equation*}

Consider the operator

\begin{equation*}
Z:l_{1}(\mathbb{I}^{d})\rightarrow S;~Z(\mu )=\left( \Psi _{0}\circ \mu
,...,\Psi _{2d}\circ \mu \right)
\end{equation*}%
and its conjugate

\begin{equation*}
Z^{\ast }:S^{\ast }\rightarrow l_{1}(\mathbb{I}^{d})^{\ast }.
\end{equation*}%
Note that by the definition of conjugate operator

\begin{equation*}
Z^{\ast }g[\mu ]=g[Z\mu ]=\sum_{q=0}^{2d}\int_{Y_{q}}g_{q}d(\Psi _{q}\circ
\mu )=\sum_{q=0}^{2d}\int_{\mathbb{I}^{d}}g_{q}\circ \Psi _{q}d\mu =\int_{%
\mathbb{I}^{d}}\left( \sum_{q=0}^{2d}g_{q}\circ \Psi _{q}\right) d\mu .
\end{equation*}%
Therefore,

\begin{equation*}
Z^{\ast }g=\sum_{q=0}^{2d}g_{q}\circ \Psi _{q}.\eqno(2.5)
\end{equation*}

Again by the above mentioned fact of Functional Analysis the operator $%
Z^{\ast }$ is surjective if and only if there exists $\epsilon >0$ such that

\begin{equation*}
\left\Vert \mu \right\Vert \leq \epsilon \left\Vert Z\mu \right\Vert
=\epsilon \sum_{q=0}^{2d}\left\Vert \Psi _{q}\circ \mu \right\Vert
\end{equation*}%
holds for all $\mu \in l_{1}(\mathbb{I}^{d})$. But we have seen above in
(2.4) that this inequality indeed holds for all $\mu \in C(\mathbb{I}%
^{d})^{\ast }$, hence for all $\mu \in l_{1}(\mathbb{I}^{d})\subset C(%
\mathbb{I}^{d})^{\ast }$. We obtain that $Z^{\ast }$ is a surjective
operator and hence its range is all of $l_{1}(\mathbb{I}^{d})^{\ast }=B(%
\mathbb{I}^{d}).$ Comparing this assertion with (2.5) gives that for any
bounded function $f\in B(\mathbb{I}^{d})$ the representation

\begin{equation*}
f=\sum_{q=0}^{2d}g_{q}\circ \Psi _{q},
\end{equation*}%
is valid for some bounded $g_{q}\in B(Y_{q}).$ Since the sets $Y_{q},$ $%
q=0,1,...,2d,$ are pairwise disjoint, we can construct a single function $g$
as in (2.2), which is well defined and bounded on $Y$. One can extend $g$ to
the whole real line in such a way that $g$ remains bounded also on $\mathbb{R%
}$. Thus for any $f\in B(\mathbb{I}^{d})$ we have the precise representation

\begin{equation*}
f(x_{1},...,x_{d})=\sum_{q=0}^{2d}g(\Psi _{q}(x_{1},...,x_{d})),\eqno(2.6)
\end{equation*}%
where $g\in B(\mathbb{R})$. If $f$ is bounded and discontinuous, then $g$
will be discontinuous as well, otherwise the right-hand side of (2.6) will
be a continuous function whereas the left-hand side is discontinuous. This
proves the second part of the theorem. It should be remarked that the method
of using $l_{1}(X)$ measures in representation of bounded functions on $X$
was introduced by Sternfeld \cite{Ste} and refined by Khavinson \cite[Ch. 1]{Kha}.

\bigskip

Now let us prove the theorem for an unbounded function $f:\mathbb{I}%
^{d}\rightarrow \mathbb{R}$. In fact, we will prove the validity of (2.1)
for any $d$-variable function $f$ defined on $\mathbb{I}^{d}$. Certainly, if
$f$ is unbounded, then $g$ must be unbounded too, otherwise the right hand
side of (2.1) would be bounded, contradicting the equality in (2.1). In \cite%
{Is1}, the second author proved a slightly different version of this part,
where not a single but $2d+1$ outer functions $g_{q}$ in the representation
formula are needed. For completeness we repeat the main details of that
proof here and show where and how all that $g_{q}$ can be replaced by a
single function $g$.

Again we use the representation formula (2.1) for continuous functions,
which has been proved above. In fact, we will use the following concise form
of (2.1)

\begin{equation*}
f(x_{1},...,x_{d})=\sum_{q=0}^{2d}g(\Psi _{q}(x_{1},...,x_{d})),\eqno(2.7)
\end{equation*}%
where $f\in C(\mathbb{I}^{d})$, $g\in C(\mathbb{R})$ and $\Psi _{q}$ are
defined above. From (2.7) we can obtain the following important property of
the family of functions $\left\{ \Psi _{q}\right\} _{q=0}^{2d}$, which we
formulate as a lemma.

\bigskip

\textbf{Lemma 2.1.} \textit{For any finite set $\{\mathbf{x}_{1},...,\mathbf{%
x}_{n}\}\subset $ $\mathbb{I}^{d}$ the system of equations}

\begin{equation*}
\sum_{j=1}^{n}\mu _{j}\delta _{\Psi _{q}(\mathbf{x}_{j})}=0,~q=0,...,2d,\eqno%
(2.8)
\end{equation*}%
\textit{with respect to $\mu _{j}$ has only a zero solution.}

\bigskip

In this lemma and in the sequel $\delta _{\mathcal{A}}$ stands for the
indicator function of a set $\mathcal{A}\subset \mathbb{R}$. That is,

\begin{equation*}
\delta _{\mathcal{A}}(t)=\left\{
\begin{array}{c}
1,\text{ if }t\in \mathcal{A} \\
0,\text{ if }t\notin \mathcal{A}%
\end{array}%
\right. .
\end{equation*}%
Note that in (2.8) $\delta _{\Psi _{q}(\mathbf{x}_{j})}$ are indicator
functions of the single point sets $\{\Psi _{q}(\mathbf{x}_{j})\}$.

\bigskip

Let us explain Eq. (2.8) in detail. We will see that it stands for a system
of certain linear equations. To show this, fix the subscript $q.$ Let the
set $\{\Psi _{q}(\mathbf{x}_{j}),$ $j=1,...,n\}$ have $s_{q}$ different
values, which we denote by $\gamma _{1}^{q},\gamma _{2}^{q},...,\gamma
_{s_{q}}^{q}.$ Then (2.8) implies that

\begin{equation*}
\sum_{j}\mu _{j}=0,
\end{equation*}%
where the sum is taken over all $j$ such that $\Psi _{q}(\mathbf{x}%
_{j})=\gamma _{k}^{q}$, $k=1,...,s_{q}$. Thus for fixed $q$ we have $s_{q}$
linear homogeneous equations in $\mu _{1},...,\mu _{n}$. The coefficients of
these equations are $0$ and $1$. By varying $q$, we will have $%
\sum_{q=0}^{2d}s_{q}$ such equations. Thus (2.8), in its expanded form,
stands for the system of these equations.

Note that not only $\left\{ \Psi _{q}\right\} _{q=0}^{2d}$, but a family of
arbitrarily given multivariate functions $\left\{ h_{1},...,h_{k}\right\} $
on any set $X$ can generate (2.8). It should be remarked that in this case
finite sets $\{\mathbf{x}_{1},...,\mathbf{x}_{n}\}$ satisfying the system of
equations (2.8) when there is a nonzero solution $(\mu _{1},...,\mu _{n})$
were exploited under the name of ``closed paths" in several works of the
second author (see, e.g., \cite{Is2,Is3,Notes}).

Let us now prove the lemma. Assume the contrary. Assume that there is a
finite set $p=\{\mathbf{x}_{1},...,\mathbf{x}_{n}\}$ in $\mathbb{I}^{d}$
such that the system of equations (2.8) has a nonzero solution $\mu =(\mu
_{1},...,\mu _{n})$. Without loss of generality we may assume that all the
numbers $\mu _{j}\neq 0$, $j=1,...,n$. Otherwise we can remove all zero
components $\mu _{j}$ from $\mu $ and the corresponding $\mathbf{x}_{j}$
(having the same index) from $p$ and consider the resulting sets $\{\mu
_{j}\}$ and $\{\mathbf{x}_{j}\}$. Consider the linear functional

\begin{equation*}
G_{p}(f)=\sum_{j=1}^{n}\mu _{j}f(\mathbf{x}_{j}),\text{ }G_{p}:C(\mathbb{I}%
^{d})\rightarrow \mathbb{R}\text{.}
\end{equation*}

This functional annihilates all sums of the form $\sum_{q=0}^{2d}g(\Psi
_{q}(x_{1},...,x_{d}))$, and hence, according to (2.7), every function $f\in
C(\mathbb{I}^{d})$. That is, $G_{p}(f)=0$ for any $f\in C(\mathbb{I}^{d})$.
On the other hand by Urysohn's lemma (see, e.g., \cite[Ch. 5]{Wil}) there
exists a continuous function $f_{0}$ with the property: $f_{0}(\mathbf{x}%
_{j})=1$ for indices $j$ such that $\mu _{j}>0$; $f_{0}(\mathbf{x}_{j})=-1$
for indices $j $ such that $\mu _{j}<0$; and $-1<f_{0}(\mathbf{x})<1$ for $%
\mathbf{x}\in \mathbb{I}^{d}\setminus p$. For this function we have $%
G_{p}(f_{0})=\sum_{j=1}^{n}\left\vert \mu _{j}\right\vert \neq 0$. The
obtained contradiction means that Eq. (2.8) has only a zero solution $(\mu
_{1},...,\mu _{n})$ for any finite subset $\{\mathbf{x}_{1},...,\mathbf{x}%
_{n}\}\subset \mathbb{I}^{d}$. The lemma has been proved.

\bigskip

Now let us return to the proof of the third part of our theorem. We want to
prove that for any function $f:\mathbb{I}^{d}\rightarrow \mathbb{R}$, the
following representation holds
\begin{equation*}
f(x_{1},...,x_{d})=\sum_{q=0}^{2d}g(\Psi _{q}(x_{1},...,x_{d})),
\end{equation*}%
where $g$ is a one-variable function depending on $f$.

Recall that we denoted ranges of $\Psi _{q}$ by $Y_{q}$ and $Y$ is the union
of $Y_{q}.$ Consider the following set
\begin{equation*}
\mathcal{L}=\{A=\{y_{0},...,y_{2d}\}:\text{if there exists }\mathbf{x}\in
\mathbb{I}^{d}\text{ s.t. }\Psi _{q}(\mathbf{x})=y_{q},~q=0,...,2d\}\eqno%
(2.9)
\end{equation*}%
Note that $\mathcal{L}$ is not a subset of $Y$. It is a set of some special
subsets of $Y.$ Each element of $\mathcal{L}$ is a set $A=\{y_{0},...,y_{2d}%
\}\subset Y$ with the property that there exists at least one point $\mathbf{%
x}\in \mathbb{I}^{d}$ such that $\Psi _{q}(\mathbf{x})=y_{q},~q=0,...,2d.$
These $\mathbf{x}$ will be called generating points for $A$.

It is almost obvious that in (2.9) for each element $A$ there exists only
one generating point $\mathbf{x}\in \mathbb{I}^{d}$. Indeed, if there were
two points $\mathbf{x}_{1}$ and $\mathbf{x}_{2}$ for a single $A$ in (2.9),
then $\Psi _{q}(\mathbf{x}_{1})=\Psi _{q}(\mathbf{x}_{2})$, $q=0,...,2d$,
and hence for the set $\left\{ \mathbf{x}_{1},\mathbf{x}_{2}\right\} \subset
\mathbb{I}^{d}$ we would have
\begin{equation*}
1\cdot \delta _{\Psi _{q}(\mathbf{x}_{1})}+(-1)\cdot \delta _{\Psi _{q}(%
\mathbf{x}_{2})}\equiv 0,~q=0,...,2d.
\end{equation*}%
That is, for such $\left\{ \mathbf{x}_{1},\mathbf{x}_{2}\right\} $ the
system of equations (2.8) would have a nonzero solution $(1;-1).$ But this
contradicts Lemma 2.1.

Since in (2.8) each $A\in \mathcal{L}$ corresponds to only one generating
point $\mathbf{x}\in \mathbb{I}^{d}$, we can define the following set
function%
\begin{equation*}
\tau :\mathcal{L}\rightarrow \mathbb{R},~\tau (A)=f(\mathbf{x}).
\end{equation*}%
Consider now a class $\mathcal{S}$ of functions of the form $%
\sum_{j=1}^{m}r_{j}\delta _{D_{j}},$ where $m$ is a positive integer, $r_{j}$
are real numbers and $D_{j}$ are elements of $\mathcal{L},~j=1,...,m.$ We
fix neither the numbers $\ m,~r_{j},$ nor the sets $D_{j}.$ Clearly, $%
\mathcal{S\ }$is a linear space. On elements of $\mathcal{S}$, we define the
linear functional
\begin{equation*}
F:\mathcal{S}\rightarrow \mathbb{R},~F\left( \sum_{j=1}^{m}r_{j}\delta
_{D_{j}}\right) =\sum_{j=1}^{m}r_{j}\tau (D_{j}).
\end{equation*}

Introduce the linear space:

\begin{equation*}
\mathcal{S}^{\prime }=\left\{ \sum_{j=1}^{m}r_{j}\delta _{\omega
_{j}}\right\} ,
\end{equation*}%
where $m\in \mathbb{N}$, $r_{j}\in \mathbb{R}$, $\omega _{j}\subset Y.$ As
above, we do not fix the parameters $m$, $r_{j}$ and $\omega _{j}.$ Note
that now we use not only the special subsets $D_{j}$ of $Y$, but all
possible subsets $\omega _{j}\subset Y$. Obviously, the space $\mathcal{S}%
^{\prime }$ is larger than $\mathcal{S}$. Consider the linear extension of $%
F $ to the space $\mathcal{S}^{\prime }$, which we denote by $F^{\prime }$.
That is, $F^{\prime }:\mathcal{S}^{\prime }\rightarrow \mathbb{R}$ and $%
F^{\prime }(z)=F(z)$ for all $z\in \mathcal{S}$.

Define the following functions:
\begin{equation*}
g_{q}:Y_{q}\rightarrow \mathbb{R},\text{ }g_{q}(y_{q}):=F^{\prime }(\delta
_{y_{q}}),\text{ }q=0,...,2d.
\end{equation*}%
Since the sets $Y_{q}$ are pairwise disjoint we can also define the single
function
\begin{equation*}
g:Y\rightarrow \mathbb{R}\text{, }g(y)=g_{q}(y),\text{ if }y\in Y_{q}\text{.}%
\eqno(2.10)
\end{equation*}%
Clearly, (2.10) correctly defines $g$ on the whole $Y$, the union of ranges
of $\Psi _{q}$.

Let now $\mathbf{x}=(x_{1},...,x_{d})$ be an arbitrary point in $\mathbb{I}%
^{d}.$ Obviously, $\mathbf{x}$ is a generating point for some set $%
A=\{y_{0},...,y_{2d}\}\in \mathcal{L}$. Thus we can write that
\begin{equation*}
f(\mathbf{x})=\tau (A)=F(\delta _{A})=F\left( \sum_{q=0}^{2d}\delta
_{y_{q}}\right) =F^{\prime }\left( \sum_{q=0}^{2d}\delta _{y_{q}}\right) =
\end{equation*}

\begin{equation*}
\sum_{q=0}^{2d}F^{\prime }(\delta
_{y_{q}})=\sum_{q=0}^{2d}g_{q}(y_{q})=\sum_{q=0}^{2d}g(y_{q})=%
\sum_{q=0}^{2d}g(\Psi _{q}(\mathbf{x})).
\end{equation*}%
This proves the theorem for all functions $f:\mathbb{I}^{d}\rightarrow
\mathbb{R}$, hence for unbounded $f$.

\bigskip

\begin{center}
{\large \textbf{3. Conclusions and remarks }}
\end{center}

The topic on the role of Kolmogorov superposition theorem in neural network
theory is still active today (see, e.g., \cite{Jorg,Mont,Sch,Shen}). The
research in this area was developed mainly in two directions. In the first
direction, the analysis was concentrated on approximative versions of
Kolmogorov's superposition theorem and obtaining corresponding results on
neural network approximation (see, e.g., \cite{GI,Is14,Notes,Kur,Kur1,MP}).
In the second direction, representation power of Kolmogorov
superposition-based neural networks were studied (see, e.g., \cite%
{Brat,Is1,Kat,Spr1,Spr2,Spr3}). Due to these second type works, there is a
perspective for a practical usage of the precise representation of
multivariate functions by Kolmogorov type neural networks.

This paper studies the Kolmogorov two hidden layer neural network model with
one-variable activation functions $\varphi $ and $g$ in the first and second
hidden layers, respectively. It shows that each multivariate function $f$
can be precisely represented by this model. All parameters of the network
except the second activation $g$ are fixed and do not depend on $f$. The
main result proves that if $f$ is continuous, then $g$ can be chosen
continuous as well. Further if $f$ is discontinuous bounded, then $g$ is
discontinuous bounded; and if $f$ is unbounded, then $g$ is unbounded.

It should be remarked that existence results and construction methods for
the universal inner function $\varphi $ were given in the papers \cite%
{Spr2,Kop,Bra,Bra1} (see Introduction). Using these methods one can easily
construct the functions $\Psi _{q}$, since all the numbers in their
definitions are explicitly known. A numerical algorithm for the parallel
computations $g_{q}$ in the representation formula (1.1) was developed in
\cite{Spr3}. Taking into account (2.2) one can apply Sprecher's method for
computation of the single outer function $g$. These tips refer to
computational aspects of the case when in Theorem 2.1 only continuous
functions are involved. A practical construction of $g$ in cases with
discontinuous bounded and unbounded functions is not yet known. For such
cases Theorem 2.1 gives only a theoretical understanding of the
representation problem. This is because for the representation of
discontinuous bounded functions we have derived (2.1) from the fact that the
range of the operator $Z^{\ast }$ is the whole space of bounded functions $B(%
\mathbb{I}^{d})$. This fact directly gives us a formula (2.1) but does not
tell how the bounded one-variable function $g$ is attained. For the
representation of unbounded functions we have used a linear extension of the
functional $F$, existence of which is based on Zorn's lemma (see, e.g., \cite%
[Ch. 3]{KF}). Application of Zorn's lemma provides no mechanism for
practical construction of such an extension. Zorn's lemma helps to assert
only its existence.

\end{document}